\documentclass{article}

\usepackage[nonatbib, final]{neurips_bdl2019}

\usepackage[utf8]{inputenc} 
\usepackage[T1]{fontenc}    
\RequirePackage{doi}
\usepackage{url}            
\usepackage{booktabs}       
\usepackage{amsfonts}       
\usepackage{amsmath}
\usepackage{nicefrac}       
\usepackage{microtype}
\usepackage{graphicx}
\usepackage{subfig}
\usepackage{enumitem}
\usepackage{wrapfig}
\usepackage{mathtools}
\usepackage{pgffor}
\usepackage{grffile}
\usepackage{caption}
\usepackage{tikz}
\usepackage{multirow}
\usepackage{subfiles}
\usepackage{varioref}
\usepackage{etoolbox}
\usepackage[title]{appendix}

\usetikzlibrary{patterns}
\usetikzlibrary{bayesnet}
\usetikzlibrary{positioning}
\usetikzlibrary{arrows,decorations.markings}
\tikzstyle{disent_latent} = [circle,pattern=north east lines, pattern color=black!20,draw=black,inner sep=1pt,
minimum size=20pt, font=\fontsize{10}{10}\selectfont, node distance=1]

\DeclareMathOperator{\expect}{\mathbb{E}}
\DeclareMathOperator*{\KL}{KL}

\DeclareMathOperator{\ELBO}{\mathcal{L}}

\title{Disentangling to Cluster: Gaussian Mixture Variational Ladder Autoencoders}

\author{
  Matthew Willetts, Stephen Roberts, Chris Holmes\\
  University of Oxford \\
  Alan Turing Institute\\
  \texttt{\{mwilletts, sroberts, cholmes\}@turing.ac.uk} \\
}

\begin{document}

\maketitle

\begin{abstract}
In clustering we normally output one cluster variable for each datapoint.
However it is not necessarily the case that there is only one way to partition a given dataset into cluster components.
For example, one could cluster objects by their colour, or by their type.
Different attributes form a hierarchy, and we could wish to cluster in any of them.
By disentangling the learnt latent representations of some dataset into different layers for different attributes we can then cluster in those latent spaces.
We call this \textit{disentangled clustering}.
Extending Variational Ladder Autoencoders (Zhao et al., 2017), we propose a clustering algorithm, VLAC, that outperforms a Gaussian Mixture DGM in cluster accuracy over digit identity on the test set of SVHN.
We also demonstrate learning clusters jointly over numerous layers of the hierarchy of latent variables for the data, and show component-wise generation from this hierarchical model.

\end{abstract}



\section{Introduction}

What do we mean when we talk of clustering a set of images?
We have data $x \in \mathcal{D}$ available at train time.
The model assigns some cluster variable $y \in 1,...,K$ to each datapoint $x$.
We score our method using some existing ground-truth label information $t \in 1,...,T$ that was not used when we applied our clustering algorithm.

But plausibly there are different ways to cluster the same set of images, so it is limiting to insist \textit{a priori} that there is only one cluster variable $y$ per datapoint and that the clustering algorithm must successfully match that $y$ to $t$ over some dataset.
Consider clustering images of digits.
While it might make sense to cluster them against the ground truth classes corresponding to what digit they represent, one could conceivably cluster based on other aspects: What colour is the digit? What colour is the background? What is the style of the typeface?

As we analyse more complex data, intuitively we can expect to find an increasing number of different aspects on which one could plausibly cluster.
To have one latent cluster variable capturing all the different aspects, the number of cluster components needed would be the product of the number of clusters for each aspect.
For example, for digits, one might need a cluster component for each combination of digit identity and colour.

Because having such a large number of cluster components would be unwieldy and unparsimonious, we are interested in outputting a set of $L$ cluster variables $\{y_\ell\}$, one of which might correspond to a particular given ground truth label and the others may capture other ways of clustering the data.
This broadened conception of clustering, which we call \textit{disentangled clustering}, requires us to learn sets of latent variables at different levels of the hierarchy of attributes so as to perform clustering over them.

So we wish for disentangled representations.
Further we want these representations to be ordered in some way.
The Variational Ladder Autoencoder (VLAE) \cite{Zhao2017_hf} provides this.
It separates out subsets of latent variables of images via the degree of computation needed to map between each layer of latent variable and the image: `high level' and `low level' aspects of the image have their associated latent variables separated from each other by how expressive the mapping is between that latent variable and the data.

We augment VLAEs so they can perform clustering at each layer, by introducing mixture distributions for each subset of latent variables.
We call this model \textit{Variational Ladder Autoencoder for Clustering} or \textit{VLAC}.
Like other deep generative models that are trained using amortised variational inference, we produce recognition networks as inference artifacts that can be applied after training to new data.
This enable us to perform \textit{test set clustering}.

We find that we can learn a clustering variable in the hierarchy that corresponds to the ground truth for SVHN \cite{SVHN}, and that we get better test set clustering accuracy than when using a single entangled layer of latent variables in a Gaussian Mixture DGM \cite{Shu2016, Nalisnick2016, Willetts2019a}.

\section{Related Work}
Our approach has some links to multi-view clustering \cite{Blum1998, Bickel2004, Xu2013, Wang2013_mvclust, Gao2015, Zhao2017_mvclust}, where like in multi-view learning one has a feature vector that is composed of distinct chunks of features each about some different aspect of that datapoint.
These subsets of features can then be each be used to produce clustering assignments.
Often the aim is to use these different sources of information to try to create the same clustering assignments \cite{Blum1998, Bickel2004}.

However, unlike multi-view clustering, we do not have access to the already-chunked feature vector that divides up the different aspects one could cluster over.
And further, whereas in multi-view clustering the different views are used to bolster one overall clustering assignment for each datapoint, here want to cluster distinctly in each learnt set of latent variables.

Various deep learning-based algorithms have been proposed for clustering, including:
Gaussian Mixture DGMs \cite{Shu2016, Nalisnick2016, Willetts2019a}, GM-VAE \cite{Dilokthanakula}, VaDE \cite{Jiang2017}, IMSAT \cite{Hu2017}, DEC \cite{Xie2016a} and ACOL-GAR \cite{Kilinc2018}.

Many recent papers on learning disentangled representations are based around achieving statistical independence between the different dimensions of the latent variables in the aggregate posterior $q(z) = \sum_{i=1}^{|\mathcal{D}|} q_\phi(z|x_i)$.
This then leads to simple generation of synthetic data by ancestral sampling, where each dimension in the learnt latent variable then controls a single (often human interpretable) aspect of the data.
Examples of this include Factor VAE \cite{Kim2018}, $\beta$-TCVAE \cite{Chen2018} and HFVAE \cite{Esmaeili2018}.
These approaches do not learn hierarchies of disentangled factors, having only one stochastic layer, and in their approach are orthogonal to the method of disentangling by degree of computation that is used by VLAEs.

Recent theoretical work has studied definitions of disentangling around symmetry groups \cite{Higgins2018} and the effect of different priors in DGMs on their posterior representations \cite{Mathieu2019}.
In \cite{Mathieu2019}, different posited varieties of disentangling (such as: the above idea of axis alignment of interpretable generative factors, learning of sparse representations, or learning to cluster) arise from different priors in DGMs when the ELBO has been augmented to contain a divergence between the aggregate posterior $q(z)$ and the prior $p(z)$, that one aims to minimise.

While our model does not include this additional divergence in our objective, our approach is an example of trying to obtain a particular variety of disentangling -- here a hierarchy of layers of variables that internally demonstrate clustering -- through the interplay of a prior and a suitable variational posterior.

Other works closely similar to ours include DGMs that attempt to learn the structure of network of latent variables, given the data.
For example, \cite{hdpvae} learns a Bayesian non-parametric model, a nested Chinese Restaurant Process \cite{Teh2006}, as the generative model for a VAE.

The work philosophically most similar to ours is \cite{multidimcluster}, where the authors aim to learn a set of clustering variables given the data, that like ours describe different aspects of the data.

Unlike these structure-learning approaches, we choose to constrain the structure of the hierarchy of latent variables in the generative model to be much simpler -- our discrete latent variables are all independent in the generative model.

Further, work in this area has often focused on learning not on raw image data but on features extracted by some pre-trained method.
A popular choice is the activations in the penultimate layer of a deep net trained on Imagenet \cite{imagenet}.
\cite{hdpvae, multidimcluster} are examples of this.
Unlike that work, our method trains on raw image data directly.
This is beneficial as it means the model has access to more information.
We are not constrained only to have access to the aspects of the data that have been picked out so as to be useful for the task under which the feature extractor was trained.
The features needed to classify an image can even be optimised by throwing away information we might care about.

For instance, in training a classifier on SVHN, one could imagine that dropping (most) colour information as quickly as possible might be worthwhile -- if in the dataset there is no correlation between colour and digit identity.
The classification training objective would be telling us to view colour information as noise on the signal we care about, and thus should play a minimal role in the embedding we get in the final layer.
Thus, we are interested in learning directly on raw image data.

\section{VLAC: Variational Ladder Autoencoders for Clustering}


\textbf{VLAEs} To gain a more expressive model over a vanilla VAE that has a single set of latent variables $z$ \cite{Kingma2013, Rezende2014}, it is natural to consider having a hierarchy of latent variables $z \rightarrow \mathbf{z} = \{z_\ell\}$ for $\ell \in 1,...,L$ each with dimensionality $d^z_\ell$.
The simplest VAE with a hierarchy of conditional stochastic variables in the generative model is the Deep Latent Gaussian Model \cite{Rezende2014}.
Here we have a Markov chain in the generative model: $p_\theta(x,\mathbf{z})=p_\theta(x|z_1)\prod_{\ell=1}^{L-1}[ p_\theta(z_\ell | z_{\ell+1})] p(z_L)$
Performing inference in this model is challenging.
The latent variables further from the data can fail to learn anything informative \cite{Sonderby2016, Zhao2017_hf}: in the worst case a single-layer VAE can train in isolation within this hierarchical model: each $p_\theta(z_{\ell}|z_{\ell+1})$ distribution can become a fixed distribution not depending on $z_{i+1}$ such that each $\KL$ divergence present in the objective between corresponding $z_\ell$ layers is driven to a local minima.
\cite{Zhao2017_hf} gives a proof of this separation for the case where the model is perfectly trained ($\KL(q_\phi(z,x)||p_\theta(x,z))=0$).

The Variational Ladder Autoencoder (VLAE) \cite{Zhao2017_hf} avoid this collapse in hierarchical VAEs.
Here we have a `flat hierarchy' in $\mathbf{z}$.
Instead of having the set of $z_\ell$ variables conditioned on each other, the prior for $\mathbf{z} = \{z_\ell\}$ is a set of independent standard Gaussians: $p(\mathbf{z}) = \prod_{\ell=1}^L \mathcal{N}(z_\ell|0,\mathcal{I})$, $|z_\ell| = {d^{z}_\ell}$.
and inside the conditional distribution $p_\theta(x|\mathbf{z})$ there is a ladder \cite{Rasmus2015, Pezeshki2016, Sonderby2016} over $z_\ell$ variables.
This separates out aspects of the data by the degree of computation needed to map between their latent representation and $x$.
Thus $p_\theta(x|\mathbf{z})$ is defined implicitly by:
\begin{align}
\tilde{z}_L &= f_L^\theta(z_L)\\
\tilde{z}_\ell &= f^\theta_\ell(z_\ell, \tilde{z}_{\ell+1}) \\
x &\sim p(x|f^\theta_0(\tilde{z}_1))
\end{align}
for $\ell \in 1,...,L-1$. 
The posterior follows a similar structure, but in reverse:
\begin{align}
h_\ell &= g^\phi_\ell(h_{\ell-1}) \label{eq:h_update}\\
z_\ell &\sim \mathcal{N}(z_\ell|\mu^\phi_\ell(h_\ell), \sigma^\phi_\ell(h_\ell)) \label{eq:qzl}
\end{align}
for $\ell \in 1,...,L$ and where $h_{0}=x$.\\


\textbf{VLAC: Variational Ladder Autoencoder for Clustering} To enable us to cluster, we alter the generative model above so we have a mixture distribution in $z_\ell$: $p_\theta(x,\mathbf{z}) \rightarrow p_\theta(x,\mathbf{z},\mathbf{y})=p(x|\mathbf{z})\prod_{\ell=1}^L p_\theta(z_\ell|y_\ell)p(y_\ell)$. $p(z_\ell|y_\ell) = \mathcal{N}(z_\ell|\mu_{y_\ell}, \sigma_{y_\ell})$ and $p(y_\ell) = \mathrm{Cat}(1/K_\ell)$.
Where $\mathbf{K}=\{ K_\ell\}$ is the vector of the dimensionalities of our discrete variables $\{y_\ell\}$.
Our variational posterior is now $q_\phi(\mathbf{z},\mathbf{y}|x)$.
We choose to factorise this as $q_\phi(\mathbf{z}|\mathbf{y},x)q_\phi(\mathbf{y}|x)$.
Each of these is a product over our $L$ layers.
$q_\phi(\mathbf{y}|x) = \prod_{\ell=1}^L q_\phi(y_\ell|x)$, $q_\phi(y_\ell|x) = \mathrm{Cat}(\pi^\phi_\ell(x))$.
$q_\phi(\mathbf{z}|\mathbf{y},x) = \prod_{\ell=1}^L q_\phi(z_\ell|x, \{y_i\}_{i\leq \ell})$, and so the new counterparts to Eqs (\ref{eq:h_update}-\ref{eq:qzl}) are:
\begin{align}
h_1 &= g^\phi_1(x) \label{eq:gm_x_update}\\
h_\ell &= g^\phi_\ell(h_{\ell -1},y_{\ell -1}) \label{eq:gm_h_update} \, \, \, \mathrm{for \, }\ell > 1\\
z_\ell &\sim \mathcal{N}(z_\ell|\mu^\phi_\ell(h_\ell, y_\ell), \sigma^\phi_\ell(h_\ell, y_\ell)) \label{eq:gm_qzl}
\end{align}

\begin{wrapfigure}{L}{0.55 \textwidth}
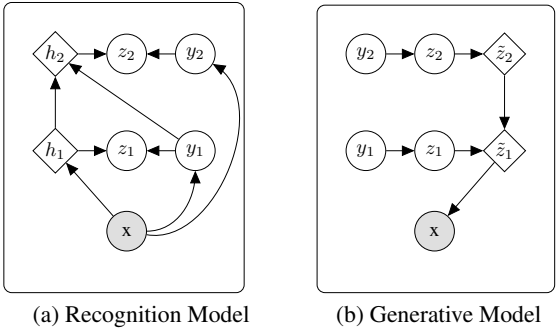

\vspace{-4mm}
\noindent\makebox[0.55 \textwidth]{%
  \subfloat[][Recognition Model]{\scalebox{0.75}{\raisebox{0ex}
{\tikz{ %
      \node[obs] (x) {x} ; %
    \node[latent, above=0.71cm of x] (z1) {$z_1$} ; %
    \node[det, left=0.5cm of z1] (h1) {$h_1$} ; %
    \node[latent, above=of z1] (z2) {$z_2$} ; %
    \node[det, left=0.5cm of z2] (h2) {$h_2$} ; %
    \node[latent, right=0.5cm of z1] (y1) {$y_1$} ; %
    \node[latent, right=0.5cm of z2] (y2) {$y_2$} ; %
    \plate[inner sep=0.5cm] {plate1} {(x) (h1) (h2) (y1) (y2) (z1) (z2)} {\scalebox{1}{}}; %
    \edge {x} {h1} ; %
    \edge {h1} {h2} ; %
    \edge {h1, y1} {z1} ; %
    \edge {y1} {h2} ; %
    \edge {h2, y2} {z2} ; %
    \draw [->] (x) to [out=0,in=-90] (y1);
    \draw [->] (x) to [out=-10,in=330] (y2);
  }}
  \label{fig:gm_inference}}}
  \hspace*{4mm}
  \subfloat[][Generative Model]{\scalebox{0.75}
{\tikz{ %
      \node[obs] (x) {x} ; %
    \node[latent, above=0.71cm of x] (z1) {$z_1$} ; %
    \node[latent, above= of z1] (z2) {$z_2$} ; %
    \node[latent, left=0.5cm of z1] (y1) {$y_1$} ; %
    \node[latent, above=of y1] (y2) {$y_2$} ; %
    \node[det, right=0.5cm of z1] (zt1) {$\tilde{z}_1$} ; %
    \node[det, right=0.5cm of z2] (zt2) {$\tilde{z}_2$} ; %
    \plate[inner sep=0.5cm] {plate1} {(x) (zt1) (zt1) (y1) (y2) (z1) (z2)} {\scalebox{1}{}}; %
    \edge {zt1} {x} ; %
    \edge {zt2} {zt1} ; %
    \edge {y1} {z1} ; %
    \edge {y2} {z2} ; %
    \edge {z1} {zt1} ; %
    \edge {z2} {zt2} ; %
  }}
 \label{fig:gm_generative}}
 }
 \caption{VLAC with $L=2$}
 \label{fig:gmvlae}
\end{wrapfigure}

See Figure \ref{fig:gmvlae} for a graphical representation of this model for $L=2$.
After training the $q_\phi(y_\ell|x)$ networks are inference artifacts that can be applied to new datapoints.

Thus the ELBO for our model is:
\begin{equation}
    \ELBO(x)=\expect_{\mathbf{z}\sim q} \log p_\theta(x|\mathbf{z}) - \sum_{\ell=1}^L \expect_{\mathbf{y} \sim q} \KL(q_\phi(z_\ell|x, \{y_i\}_{i\leq \ell}) || p(z_\ell|y_\ell)) - \sum_{\ell=1}^L \KL(q_\phi(y_\ell|x) || p(y_\ell))
\end{equation}
\setlength{\intextsep}{2pt}%

If all $K_\ell=1$ then VLAC reduces to a VLAE.
It is not necessary to have $K_\ell>1$ for all layers in VLAC.

\textbf{Evaluation Metric} Following \cite{Jiang2017, Yang2016, Kilinc2018} we use cluster accuracy (ACC), also known as cluster purity, to evaulate our models:
\begin{equation}
    \mathrm{ACC} = \max_{P \in \mathcal{P}}\frac{\sum_{i=1}^{|\mathcal{D}|}\mathbb{I}[t_i=P y_i]}{|\mathcal{D}|}
\end{equation}
where $P$ is a $T \times K$ rectangular permutation matrix that attributes each $y$ to a ground truth class $t$.

\section{Experiments}

We trained our model with $L=4$ and convolutional $g$ and deconvolutional $f$ networks.
$p(x|f^\theta_0(\tilde{z}_1))$ was a Gaussian distribution with fixed variance.
For full implementation details, see the code at\footnote{https://github.com/MatthewWilletts/VLAC}.
We used CONCRETE sampling/the Gumbel-Softmax Trick \cite{Maddison2016, Jang2016} to estimate stochastically the expectations over discrete variables, rather than exactly marginalise them out.
This avoids us having to calculate numerous forward passes through the model.

We apply our model to SVHN \cite{SVHN}, as it gives variations in style of one type of object while also having distinct ground-truth class structure (here $t$ indexes digit identity) that we can benchmark against.

When running our implementation of a VLAE over SVHN we observed that the 3rd layer was associated most clearly with variation in digit identity.
In our experiments we ran VLAC with $\mathbf{K}=\mathbf{K^{\mathrm{one}}} =[1,1,50,1]$ and $\mathbf{K^{\mathrm{two}}} =[1,5,50,1]$.
We evaluate the cluster accuracy of $y_3$ over the test set, taking as our predictions the argmax of the posterior $q_\phi(y_3|x)$.

In addition to published baseline results, we also compared against a single-$z$-layer Gaussian mixture DGM \cite{Shu2016, Nalisnick2016} with an encoder-decoder structure matching that of the sub-networks needed for the 3rd layer of VLAC with $K_3 = 50$.

\begin{figure}[h!]
\centering
\makebox[\textwidth]{
    \subfloat[$\ell=2$]{\includegraphics[height=2.5cm]{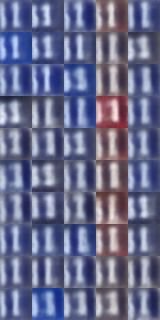}}
    \hspace{2mm}
    \subfloat[$\ell=3$]{\includegraphics[height=2.5cm]{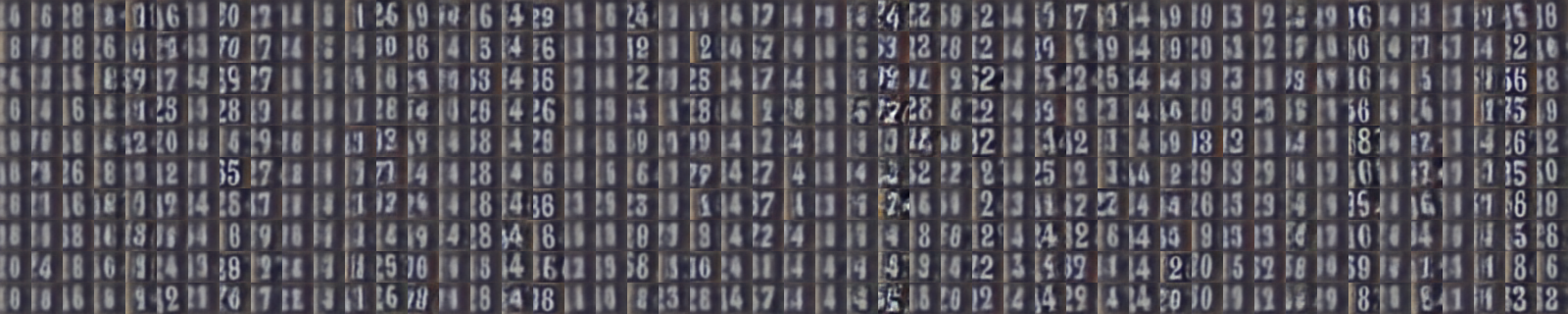}}
}
     \caption{Decoder means when sampling from VLAC with $\mathbf{K^{\mathrm{two}}}$ from layers with $K_\ell>1$. Sample from $p_\theta(z_\ell|y_\ell)$ for each cluster component $y_\ell$ for one layer while keeping all other layers fixed with one sample from their priors. One column per cluster component.}
     \label{fig:manip_class}
     \vspace{-3mm}
\end{figure}

We also perform class-conditional generation from the layers with $K_\ell>1$, sampling from each cluster component $p(z_\ell|y_\ell$).
See Figure \ref{fig:manip_class}.


\begin{table}[t!]
\vspace{1em}
  \caption{Test set cluster accuracy on SVHN for our approach and baselines.}
  \label{table:results}
  \centering
  \begin{tabular}{rc}
    \toprule
    Model & Cluster Accuracy \\
    \midrule
    \midrule
    VLAC with $\mathbf{K^{\mathrm{one}}}$   & $0.351 \pm 0.038$\\
    VLAC with $\mathbf{K^{\mathrm{two}}}$   & $0.378 \pm 0.022$\\
    \midrule
    Equivalent GM-DGM with $|K|=50$            & $0.252 \pm 0.004$\\
    \midrule
    IMSAT \cite{Hu2017}                      & $0.573 \pm 0.040$\\
    DEC \cite{Xie2016a}                       & $0.560 \pm 0.016$\\
    ACOL-GAR \cite{Kilinc2018}               & $0.768 \pm 0.013$\\
    \bottomrule
  \end{tabular}
\end{table}

\begin{figure}[h!]
\centering
\makebox[\textwidth]{
    \subfloat[$\ell=1$]{\includegraphics[height=2.5cm]{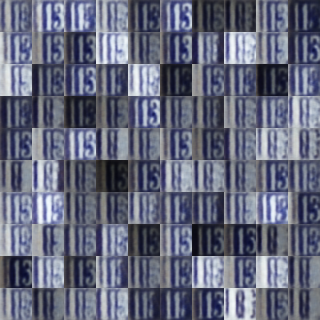}}
    \hspace{1mm}
    \subfloat[$\ell=2$]{\includegraphics[height=2.5cm]{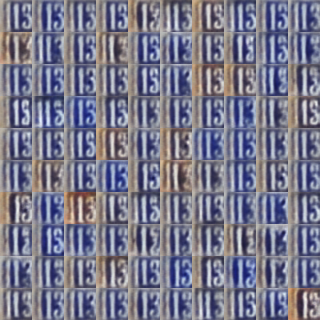}}
    \hspace{1mm}
    \subfloat[$\ell=3$]{\includegraphics[height=2.5cm]{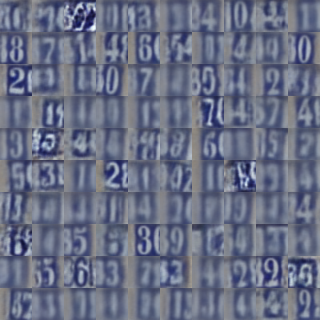}}
    \hspace{1mm}
    \subfloat[$\ell=4$]{\includegraphics[height=2.5cm]{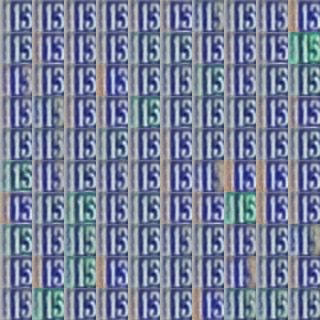}}
}
     \caption{Decoder means when resampling the latent representation of a datapoint in each layer of VLAC with $\mathbf{K^{\mathrm{two}}}$. For each layer, sample from the marginal $p_\theta(z_\ell)=\sum_i^{K_\ell}p_\theta(z_\ell|y^i_\ell)p(y^i_\ell)$ while keeping all other layers fixed.}
     \label{fig:manip_layerwise}
    \vspace{-2mm}
\end{figure}

\section{Dicussion}
Our model does not achieve state of the art clustering for SVHN.
However, we can see that from the $\mathbf{K^{\mathrm{one}}}$ results that clustering inside a ladder of stochastic variables is better than an equivalent GM-DGM baseline.
And as VLAC with $\mathbf{K^{\mathrm{two}}}$ gets better test set accuracy than VLAC with $\mathbf{K^{\mathrm{one}}}$, we see that having a hierarchy of clusters increases performance further still.

From Figures \ref{fig:manip_layerwise} $\&$ \ref{fig:manip_class} we can see that our model does separate out class information: class variation is mostly associated with one layer (as in a vanilla VLAE) and cluster components within that layer generally correspond to particular ground truth classes.
Figure \ref{fig:manip_class}a shows that the model has also discovered clusters describing the colour temperature of the image.
Overall we are pleased to have demonstrated the benefits of clustering in disentangled spaces, and hope that this inspires more research both into how to cluster datasets over different aspects of the data and how disentangling can be used to improve performance of various classical machine learning tasks when working with images.

\newpage
\section{Acknowledgements}
We thank Alexander Camuto for his helpful insights.

\bibliographystyle{unsrt}
\bibliography{references.bib}\vspace{0.75in}

\begin{thebibliography}{10}

\bibitem{Zhao2017_hf}
Shengjia Zhao, Jiaming Song, and Stefano Ermon.
\newblock {Learning Hierarchical Features from Generative Models}.
\newblock In {\em ICML}, 2017.

\bibitem{SVHN}
Yuval Netzer.
\newblock {Reading Digits in Natural Images with Unsupervised Feature Learning
  Yuval}.
\newblock In {\em NeurIPS Deep Learning and Unsupervised Feature Learning
  Workshop}, 2011.

\bibitem{Shu2016}
Rui Shu.
\newblock {Gaussian Mixture VAE: Lessons in Variational Inference, Generative
  Models, and Deep Nets}, 2016.

\bibitem{Nalisnick2016}
Eric Nalisnick, Lars Hertel, and Padhraic Smyth.
\newblock {Approximate Inference for Deep Latent Gaussian Mixtures}.
\newblock In {\em NeurIPS Bayesian Deep Learning Workshop}, 2016.

\bibitem{Willetts2019a}
Matthew Willetts, Stephen~J Roberts, and Christopher~C Holmes.
\newblock {Semi-Unsupervised Learning with Deep Generative Models: Clustering
  and Classifying using Ultra-Sparse Labels}.
\newblock {\em CoRR}, 2019.

\bibitem{Blum1998}
Avrim Blum and Tom Mitchell.
\newblock {Combining labeled and unlabeled data with co-training}.
\newblock In {\em Proceedings of the 11th Conference on Computational Larning
  Theory}, pages 92--100, 1998.

\bibitem{Bickel2004}
Steffen Bickel and Tobias Scheffer.
\newblock {Multi-View Clustering}.
\newblock In {\em ICDM}, 2004.

\bibitem{Xu2013}
Chang Xu and Chao Xu.
\newblock {A Survey on Multi-view Learning}.
\newblock {\em CoRR}, 2013.

\bibitem{Wang2013_mvclust}
Hua Wang and Heng Huang.
\newblock {Multi-View Clustering and Feature Learning via Structured Sparsity}.
\newblock In {\em ICML}, 2013.

\bibitem{Gao2015}
Hongchang Gao.
\newblock {Multi-View Subspace Clustering}.
\newblock In {\em ICCV}, 2015.

\bibitem{Zhao2017_mvclust}
Handong Zhao, Zhengming Ding, and Yun Fu.
\newblock {Multi-view clustering via deep matrix factorization}.
\newblock In {\em AAAI}, 2017.

\bibitem{Dilokthanakula}
Nat Dilokthanakul, Pedro A~M Mediano, Marta Garnelo, Matthew C~H Lee, Hugh
  Salimbeni, Kai Arulkumaran, and Murray Shanahan.
\newblock {Deep Unsupervised Clustering with Gaussian Mixture VAE}.
\newblock {\em CoRR}, 2017.

\bibitem{Jiang2017}
Zhuxi Jiang, Yin Zheng, Huachun Tan, Bangsheng Tang, and Hanning Zhou.
\newblock {Variational Deep Embedding: An Unsupervised and Generative Approach
  to Clustering}.
\newblock In {\em IJCAI}, 2017.

\bibitem{Hu2017}
Weihua Hu, Takeru Miyato, Seiya Tokui, Eiichi Matsumoto, and Masashi Sugiyama.
\newblock {Learning Discrete Representations via Information Maximizing
  Self-Augmented Training}.
\newblock In {\em ICML}, 2017.

\bibitem{Xie2016a}
Junyuan Xie, Ross Girshick, and Ali Farhadi.
\newblock {Unsupervised Deep Embedding for Clustering Analysis}.
\newblock {\em ICML}, 2016.

\bibitem{Kilinc2018}
Ozsel Kilinc and Ismail Uysal.
\newblock {Learning Latent Representations in Neural Networks for Clustering
  Through Pseudo Supervision and Graph-based activity Regularization}.
\newblock In {\em ICLR}, 2018.

\bibitem{Kim2018}
Hyunjik Kim and Andriy Mnih.
\newblock {Disentangling by Factorising}.
\newblock In {\em NeurIPS}, 2018.

\bibitem{Chen2018}
Ricky T~Q Chen, Xuechen Li, Roger Grosse, and David Duvenaud.
\newblock {Isolating Sources of Disentanglement in Variational Autoencoders}.
\newblock In {\em NeurIPS}, 2018.

\bibitem{Esmaeili2018}
Babak Esmaeili, Hao Wu, Sarthak Jain, Alican Bozkurt, N~Siddharth, Brooks
  Paige, Dana~H Brooks, Jennifer Dy, and Jan-Willem van~de Meent.
\newblock {Structured Disentangled Representations}.
\newblock In {\em AISTATS}, 2019.

\bibitem{Higgins2018}
Irina Higgins, David Amos, David Pfau, Sebastien Racaniere, Loic Matthey,
  Danilo Rezende, and Alexander~Lerchner Deepmind.
\newblock {Towards a Definition of Disentangled Representations}.
\newblock {\em CoRR}, 2018.

\bibitem{Mathieu2019}
Emile Mathieu, Tom Rainforth, N.~Siddharth, and Yee~Whye Teh.
\newblock {Disentangling Disentanglement in Variational Autoencoders}.
\newblock In {\em ICML}, 2019.

\bibitem{hdpvae}
Prasoon Goyal, Zhiting Hu, Xiaodan Liang, Chenyu Wang, and Eric~P Xing.
\newblock {Nonparametric Variational Auto-Encoders for Hierarchical
  Representation Learning}.
\newblock In {\em Proceedings of the IEEE International Conference on Computer
  Vision}, volume 2017-Octob, pages 5104--5112, 2017.

\bibitem{Teh2006}
Yee~Whye Teh, Michael~I Jordan, Matthew~J Beal, and David~M Blei.
\newblock {Hierarchical Dirichlet processes}.
\newblock {\em Journal of the American Statistical Association},
  101(476):1566--1581, 2006.

\bibitem{multidimcluster}
Xiaopeng Li, Zhourong Chen, Leonard~K.M. Poon, and Nevin~L Zhang.
\newblock {Learning latent superstructures in variational autoencoders for deep
  multidimensional clustering}.
\newblock In {\em ICLR}, 2019.

\bibitem{imagenet}
J~Deng, W~Dong, R~Socher, L.-J. Li, K~Li, and L~Fei-Fei.
\newblock {ImageNet: A Large-Scale Hierarchical Image Database}.
\newblock In {\em CVPR}, 2009.

\bibitem{Kingma2013}
Diederik~P Kingma and Max Welling.
\newblock {Auto-Encoding Variational Bayes}.
\newblock In {\em NeurIPS}, 2013.

\bibitem{Rezende2014}
Danilo~Jimenez Rezende, Shakir Mohamed, and Daan Wierstra.
\newblock {Stochastic Backpropagation and Approximate Inference in Deep
  Generative Models}.
\newblock In {\em ICML}, 2014.

\bibitem{Sonderby2016}
Casper~Kaae S{\o}nderby, Tapani Raiko, Lars Maal{\o}e, Søren~Kaae S{\o}nderby,
  and Ole Winther.
\newblock {Ladder Variational Autoencoders}.
\newblock In {\em NeurIPS}, 2016.

\bibitem{Rasmus2015}
Antti Rasmus, Harri Valpola, Mikko Honkala, Mathias Berglund, and Tapani Raiko.
\newblock {Semi-Supervised Learning with Ladder Networks}.
\newblock In {\em NeurIPS}, 2015.

\bibitem{Pezeshki2016}
Mohammad Pezeshki, Linxi Fan, Philemon Brakel, Aaron Courville, and Yoshua
  Bengio.
\newblock {Deconstructing the ladder network architecture}.
\newblock In {\em ICML}, 2016.

\bibitem{Yang2016}
Jianwei Yang, Devi Parikh, Dhruv Batra, and Virginia Tech.
\newblock {Joint Unsupervised Learning of Deep Representations and Image
  Clusters}.
\newblock In {\em CVPR}, 2016.

\bibitem{Maddison2016}
Chris~J Maddison, Andriy Mnih, and Yee~Whye Teh.
\newblock {The Concrete Distribution: A Continuous Relaxation of Discrete
  Random Variables}.
\newblock In {\em ICLR}, 2017.

\bibitem{Jang2016}
Eric Jang, Shixiang Gu, and Ben Poole.
\newblock {Categorical Reparameterization with Gumbel-Softmax}.
\newblock In {\em ICLR}, 2017.

\end{thebibliography}

\newpage
\appendix
\section{Sampling from the prior of our GM-DGM}

\begin{figure}[h!]
\centering
\makebox[\textwidth]{
    \includegraphics[height=3cm]{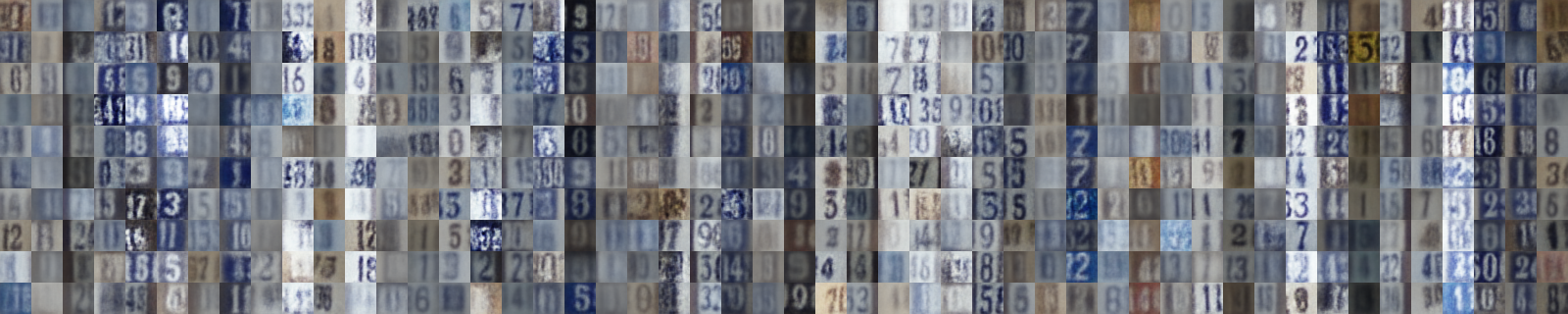}}
     \caption{Decoder means when sampling from the latent layer of GM-DGM with $|K|=50$, sampling from $p_\theta(z|y)$ for each cluster component $y$.}
     \label{fig:manip_class_gm_dgm}
\end{figure}

Comparing Figure \ref{fig:manip_class_gm_dgm} to Figure \ref{fig:manip_class}, we see that the GM-DGM encoding in $z$ entangles class, colour and style information, unlike VLAC.

\end{document}